\documentclass[letterpaper, 10 pt, conference]{ieeeconf}
\IEEEoverridecommandlockouts
\usepackage{cite}
\usepackage{amsmath,amssymb,amsfonts}
\usepackage{graphicx}
\usepackage{textcomp}
\usepackage{xcolor}
\usepackage{url}
\usepackage{tablefootnote}

\usepackage{algorithm}
\usepackage{algpseudocode}
\usepackage{subcaption}
\usepackage{multirow}
\usepackage{booktabs}
\usepackage{makecell}
\usepackage{booktabs}
\usepackage{pifont}
\usepackage{hyperref}

\title{\LARGE \bf
Unveiling the Surprising Efficacy of Navigation Understanding in End-to-End Autonomous Driving
}

\author{Zhihua Hua$^{1,3}$, Junli Wang$^{3,4}$, Pengfei Li$^{3}$, Qihao Jin$^{1,3}$, Bo Zhang$^{2}$, Kehua Sheng$^{2}$, Yilun Chen$^{3}$, \\Zhongxue Gan$^{1*}$, and Wenchao Ding$^{1*}$
\thanks{This work was supported in part by the National Natural Science Foundation of China (NSFC) under Grant 62403142, and in part by the Science and Technology Commission of Shanghai Municipality under Grant 24511103100.}
\thanks{$^{1}$ College of Intelligent Robotics and Advanced Manufacturing, Fudan University, China \{\texttt{zhhua24\}@m.fudan.edu.cn}, \{\texttt{ganzhongxue, dingwenchao\}@fudan.edu.cn}}
\thanks{$^{2}$ Didi Chuxing}
\thanks{$^{3}$ Institute for AI Industry Research (AIR), Tsinghua University}
\thanks{$^{4}$ Institute of Automation, Chinese Academy of Sciences}
\thanks{$^{*}$ Corresponding authors: Wenchao Ding and Zhongxue Gan}}

\begin{document}

\maketitle
 
\begin{abstract}
Global navigation information and local scene understanding are two crucial components of autonomous driving systems. However, our experimental results indicate that many end-to-end autonomous driving systems tend to over-rely on local scene understanding while failing to utilize global navigation information. These systems exhibit weak correlation between their planning capabilities and navigation input, and struggle to perform navigation-following in complex scenarios. To overcome this limitation, we propose the Sequential Navigation Guidance (SNG) framework, an efficient representation of global navigation information based on real-world navigation patterns. The SNG encompasses both navigation paths for constraining long-term trajectories and turn-by-turn (TBT) information for real-time decision-making logic. We constructed the SNG-QA dataset, a visual question answering (VQA) dataset based on SNG that aligns global and local planning. Additionally, we introduce an efficient model SNG-VLA that fuses local planning with global planning. The SNG-VLA achieves state-of-the-art performance through precise navigation information modeling without requiring auxiliary loss functions from perception tasks. Project page: \href{https://fudan-magic-lab.github.io/SNG-VLA-web/}{SNG-VLA}

\end{abstract}

\section{Introduction}

In recent years, end-to-end autonomous driving systems have garnered significant attention from researchers~\cite{iros-end-to-end, multi-end-to-end}. 
The end-to-end paradigm simplifies traditional modular systems, is better suited to data-driven training approaches, and demonstrates enhanced generalization performance~\cite{end-to-endautonomous}.

\begin{figure}[t]
    \centering
    \includegraphics[width=0.5\textwidth]{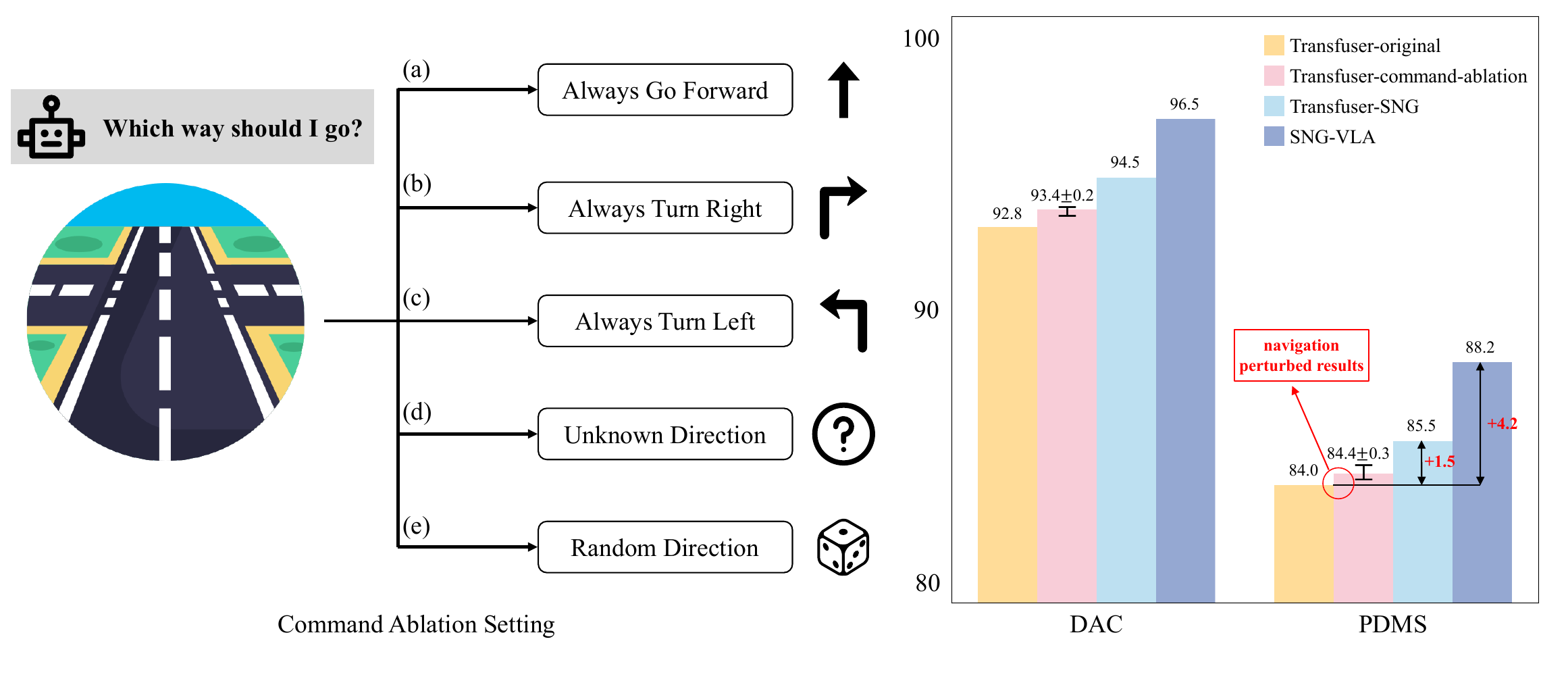}
    \caption{
    We demonstrate the impact of introducing perturbations to the driving command and the effectiveness of the SNG. The command ablation experiments comprise five distinct experimental settings, with Transfuser-command-ablation representing the statistical results across all five experiments. Detailed data are presented in Table~\ref{tab:cmd_ablation}. The utilization of random or erroneous navigation information demonstrates minimal impact on model performance. SNG exhibits significant performance improvement over the baseline approach.}
    \vspace{-2mm}
    \label{fig:teaser}
\end{figure}

Global Navigation information plays a pivotal role in end-to-end autonomous driving systems~\cite{navigation_im}, providing essential directional references for trajectory planning. Unlike prediction~\cite{multimode-prediction}, which generates multimodal trajectory forecasts, planning requires explicit navigation inputs to produce deterministic driving trajectories~\cite{vad, stp3}. Despite the critical role of global navigation information in end-to-end autonomous driving systems, we have made a surprising observation: 
\textit{removing or corrupting navigation information in existing end-to-end driving methods minimally affects planning performance and, in some cases, even improves performance}. For instance, as illustrated in Fig.~\ref{fig:teaser}, our experiments with the Transfuser~\cite{transfuser} on the NAVSIM~\cite{navsim} benchmark demonstrate that complete removal of navigation information paradoxically yields superior results. This phenomenon contradicts basic driving logic, as one would anticipate a substantial decline in planner performance when explicit navigation information is absent. This unexpected outcome raises a critical question: \textit{Do current end-to-end autonomous driving systems truly understand and utilize global navigation information?}

\begin{figure}[t]
    \centering
    \includegraphics[width=0.5\textwidth]{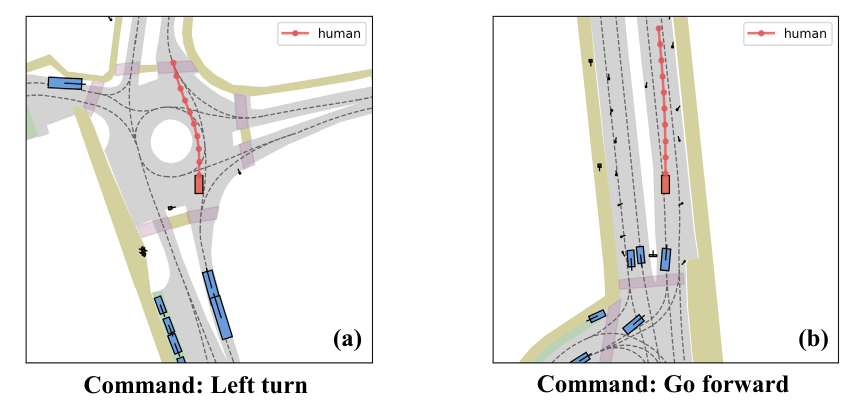}
    \caption{
    We demonstrate erroneous examples in the annotation process of driving commands. (a) Incorrect annotation occurs in roundabouts due to longitudinal displacement. (b) The annotation information fails to comprehend lane-changing behavior, resulting in causal confusion.}
    \vspace{-2mm}
    \label{fig:second}
\end{figure}

Our answer is unequivocally negative. Current research~\cite{paradrive, drivelm, genad} predominantly employs driving commands (such as ``Turn Left", ``Go Forward", ``Turn Right", ``None") to represent global navigation information, utilizing one-hot encoding to discretize driving behaviors into finite categories. However, as shown in Fig.~\ref{fig:second}, this approach exhibits the following limitations: (1) the annotation process relies on a fixed temporal horizon or spatial intervals~\cite{vad, navsim}, which can lead to ambiguous interpretations in complex scenarios. In the roundabout, the significant lateral displacement of the vehicle going forward caused the driving command to be incorrectly labeled as a ``Turn Left".
(2) this representation suffers from oversimplification. In beyond visual range (BVR) scenarios~\cite{peng2025navigscene}, a vehicle must change lanes in advance to execute a turn at a distant intersection. However, as the global navigation command is labeled as ``Go Forward", it creates causal confusion in the model when encountering lane-changing behaviors present in expert trajectories. Consequently, numerous end-to-end autonomous driving systems fail to effectively utilize navigation information, and their performance likely stems from overfitting to specific input channels~\cite{ad-mlp, bevplaner}.

To address these limitations and enhance the navigation semantic comprehension capabilities of end-to-end autonomous driving systems, we propose a novel paradigm of Sequential Navigation Guidance (SNG), inspired by real-world navigation patterns~\cite{google-map}. The SNG effectively represents navigation information by integrating static global path planning with dynamic high-level guidance:  (1) Navigation Path: A predefined trajectory segment extracted from the global path, serving as a reference line for planning; (2) Real-time Turn-by-Turn (TBT) Information: A comprehensive set of high-level guidance cues comprising current driving actions with associated distance and time estimations, future actions, and corresponding supplementary actions, which collectively inform the planning process. Notably, both types of information can be conveniently acquired through navigation APIs and are readily available off-the-shelf in practical deployment.

To further align local planning with global planning, we propose SNG-QA, which imposes additional constraints on local planning within both the reasoning and action spaces. The SNG-QA dataset comprises 100K QA pairs that decompose the reasoning process into hierarchical planning components. The local planning reasoning process integrates both global planning outcomes and local scene understanding. We utilize NAVSIM~\cite{navsim} and its annotations to construct the SNG-QA pairs according to predefined task formats. We introduce an efficient model SNG-VLA. The model employs multimodal fusion encoder and a unified transformer backbone, which can efficiently handle the constraints between global navigation information and local scene inputs. The model autoregressively generates textual reasoning and planning trajectories.

The proposed method demonstrates remarkable efficacy in modeling global navigation information, offering a plug-and-play solution that significantly enhances the planning capabilities of end-to-end autonomous driving systems. Our model also achieves state-of-the-art performance on both the Bench2Drive~\cite{bench2drive} a closed-loop benchmark based on Carla~\cite{carla} and the NAVSIM~\cite{navsim} a real-world evaluation benchmark. Our contributions can be summarized as follows:

\begin{enumerate}[label=\arabic*., leftmargin=2em]
    \item To address the limitations of current global navigation representation, we propose a novel Sequential Navigation Guidance approach to structure global navigation information, offering both long-term trajectory constraints and real-time decision-making logic.
    \item We propose the SNG-QA dataset, which partitions the reasoning process into global and local planning components to enhance local planning rationality and ensure consistency between local and global planning.
    \item We develop an effective model that, without incorporating auxiliary tasks and utilizing precise navigation information, achieves state-of-the-art (SOTA) performance in both Bench2Drive and NAVSIM benchmarks.
\end{enumerate}

\begin{figure*}[t]
    \centering
    \includegraphics[width=\textwidth]{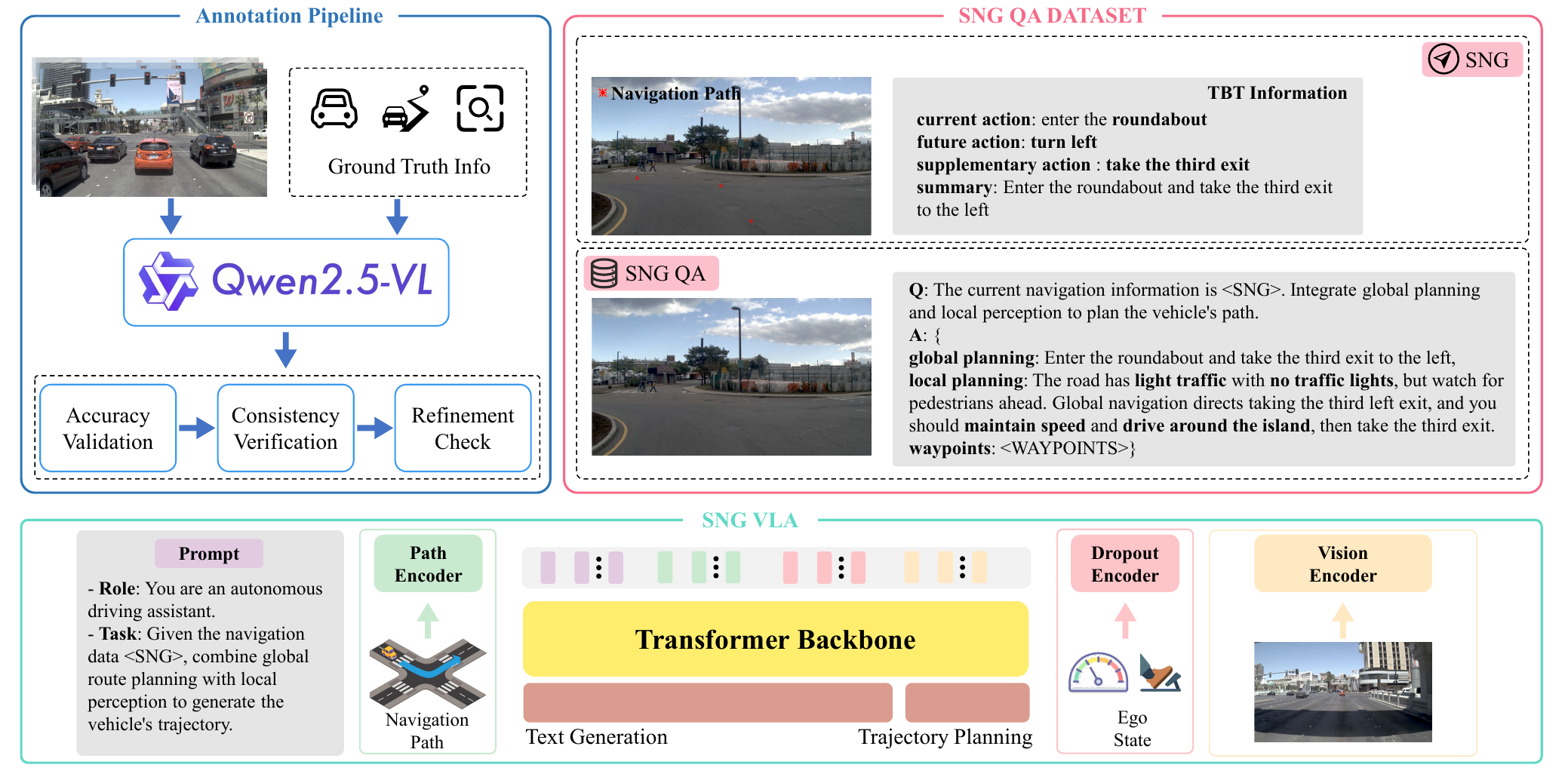}
    \caption{Overview of our pipeline. Sequential navigation guidance is consists of navigation path and TBT information. SNG-QA comprises three subtasks: global planning, local planning, and trajectory planning. The architecture of our model is divided into two parts: the multimodal feature fusion encoder and the unified transformer backbone.}
    \label{fig:pipeline}
\end{figure*}

\section{Related Work}
\label{sec:related_work}
\subsection{End-to-end autonomous driving}
\label{sec:work_pp}

Traditional autonomous driving systems are often composed of multiple modular components~\cite{bevformer,traj-plan}, whereas end-to-end autonomous driving enables a direct mapping from raw sensor data to planning trajectories~\cite{uniad, mp3}. Most methods~\cite{tcp, driveadapter} and dataset~\cite{nuscenes} utilize driving commands as global navigation information.
UniAD~\cite{uniad} has significantly enhanced the performance of autonomous driving systems by integrating multiple modules into an end-to-end framework. VAD~\cite{vad} employs a fully vectorized approach to model driving scenarios, ensuring planning safety while improving operational efficiency. BEVPlanner~\cite{bevplaner} transforms sensor inputs into BEV (Bird's Eye View) features, serving as an intermediate representation within the end-to-end architecture. GenAD~\cite{genad} introduces a novel generative framework that aids planning tasks by predicting the dynamic interactions between the ego vehicle and the environment. However, driving command constrains the practical performance of these methods.

\subsection{Multimodal large models for planning}

Multimodal large models facilitate seamless interaction and understanding across diverse data types, driving transformative innovations in fields such as natural language processing and beyond~\cite{gpt4}. Given the necessity of processing sensor data from multiple modalities and performing joint planning in autonomous driving systems, multimodal large models naturally serve as an effective backbone for such systems. DriveGPT4~\cite{drivegpt4} leverages multimodal large models to process multi-frame video and text inputs, simultaneously outputting reasoning processes during planning, thereby significantly enhancing the interpretability and interactivity of autonomous driving systems. LMDrive~\cite{lmdrive} unifies multimodal sensor data into a textual feature space, greatly improving the interactivity of autonomous driving systems and demonstrating exceptional performance in CARLA~\cite{carla} closed-loop evaluation. 
Therefore, we propose an efficient pipeline based on a multimodal large model, capable of processing data from various sensor modalities and performing end-to-end planning tasks.

\subsection{Navigation information for planning}
\label{sec:work_navi}

Navigation information plays a critical role in autonomous driving planning. Current end-to-end benchmarks primarily rely on driving commands as navigation inputs. In real-world datasets\cite{nuscenes, nuplan, waymo}, driving commands (e.g., ``Turn Left") are implicitly inferred from expert trajectories to model navigation information. Although simulators~\cite{carla, bench2drive} provide waypoints between the current position and the target location, they still use discrete driving commands as the primary input. Several methods, such as UniAD\cite{uniad} and BEVPlanner\cite{bevplaner}, embed navigation commands into latent spaces as additional model inputs. ST-P3\cite{stp3} samples multiple trajectories and filters them based on geometric features aligned with driving commands, while VAD\cite{vad} generates results for all command categories and selects the corresponding trajectory as the final output. TCP\cite{tcp} and TransFuser\cite{navsim, transfuser} concatenate driving commands with ego states as conditional inputs. However, relying solely on driving commands to model navigation information leads to intent ambiguity and deviations from real-world scenarios. To address these limitations, we propose integrating TBT (Turn-by-Turn) instructions and navigation paths to model more accurate navigation information, thereby improving planning rationality, safety, and human-like interaction.




\section{Methods}
\label{sec:methods}

The pipeline of our method is shown in Fig.~\ref{fig:pipeline}. Specifically, we first introduce the modeling approaches for sequential navigation guidance in Section~\ref{sec:navigation_model}. Subsequently, we present the construction methodology of SNG-QA in Section~\ref{sec:sng-qa}. In Section~\ref{sec:architrcture}, we detail our model architecture, which comprises a multimodal feature fusion encoder described in Section~\ref{sec:scene_rep} and a transformer-based decoder outlined in Section~\ref{sec:transformer_decoder}.

\subsection{Modeling Global Navigation Information}
\label{sec:navigation_model}

In real-world driving scenarios, most driving behaviors are guided by specific navigation information, often facilitated by tools such as Google Maps~\cite{google-map}. Navigation information typically comprises two key components: a pre-planned global route $R$, generated using A*, and real-time turn-by-turn (TBT) information $I$. The global route, when transformed from the world coordinate system to the vehicle coordinate system, serves as a reference line for the vehicle's direction of driving. Simultaneously, TBT information, which includes high-level textual prompts, provides immediate guidance for local maneuvers. 

We construct SNG by integrating the navigation path and turn-by-turn (TBT) information, as illustrated in Fig.~\ref{fig:pipeline}. 
Specifically, we select road centerlines within a 40m range ahead of each vehicle as references and sampled them to generate navigation path $P=\{(\hat{x}_1, \hat{y}_1), (\hat{x}_2, \hat{y}_2),\dots,(\hat{x}_{N_p}, \hat{y}_{N_p})\}$, $N_p$ represents the number of navigation points. To simulate real-world localization errors and mitigate the influence of privileged information in navigation paths on model performance, we introduced substantial noise to the sampled navigation paths.

The TBT information includes current driving actions with associated distance and time estimations, future actions, and future supplementary action. Drawing from real-world navigation systems, we categorize driving actions into eight distinct types: turn left, turn right, execute U-turn, proceed straight, keep left, keep right, enter roundabout, and none. We employ the vehicle's route and camera data feed as inputs to VLM for predicting both current and future driving actions. The duration of each current action is calculated based on the future trajectory and instantaneous vehicle speed. We define nine categories of supplementary actions—including entering highways, tunnels, right-turn lanes, left-turn lanes, etc.—to provide contextual information for future action prediction. When the VLM identifies ambiguity in a future action description, it incorporates an appropriate supplementary action for disambiguation. The entire annotation framework is implemented using Qwen2.5 VL 72B\cite{qwen2}.

\subsection{SNG-QA dataset}
\label{sec:sng-qa}
High-quality question-answering (QA) datasets are crucial for enhancing Vision-Language-Action (VLA) models' scene understanding and reasoning capabilities~\cite{drivelm,zhou2025autovla}. Driving behavior reasoning represents a key task in visual question answering for autonomous driving, playing a vital role in ensuring that models correctly interpret scenes and formulate reasonable planning strategies. However, existing VQA datasets~\cite{drivelm,renz2025simlingo} typically rely on local perception and expert trajectories for annotating driving behavior inferences, thereby neglecting the role of global navigation information in the annotation process. For instance, in scenarios such as beyond visual range (BVR)  lane changes or roundabout exit selections, models struggle to accurately interpret expert trajectory intentions when relying solely on local perception.

To address this limitation, we developed an automated annotation workflow based on Qwen 2.5 VL 72B\cite{qwen2} that divides driving behavior reasoning into three stages: global navigation information summarization, local planning, and trajectory point generation. The model first generates corresponding summaries based on the input SNG. To ensure that local planning fully comprehends both global navigation information and local scene understanding, we employ global navigation information and object detection labels as prompts to guide the model in generating causal explanations for local planning decisions. To guarantee the quality of the generated textual explanations, we implemented a three-stage validation process encompassing accuracy verification, consistency validation, and language refinement. Based on NAVSIM\cite{navsim}, we constructed an inference annotation dataset comprising approximately 100,000 samples.

\begin{table*}[t!]  
\setlength{\tabcolsep}{5pt}
    \centering
    \small  
    \bgroup
    \renewcommand{\arraystretch}{1.1}  
    \begin{tabular}{l|c|c|cc|ccc|c}
    \toprule[1pt]
    Method & Input & Navigation & NC$\uparrow$ & DAC$\uparrow$ & TTC$\uparrow$ & Comf.$\uparrow$ & EP$\uparrow$ & PDMS$\uparrow$ \\
    \cmidrule(lr){1-9}
    UniAD~\cite{uniad} & C \& L & CM & 97.8 & 91.9 & 92.9 & \textbf{100} & 78.8 & 83.4 \\
    PARA-Drive~\cite{paradrive} & C \& L & CM & 97.9 & 92.4 & 93.0 & 99.8 & 79.3 & 84.0 \\
    LTF~\cite{transfuser}& C \& L & CM  & 97.4 & 92.8 & 92.4 & \textbf{100} & 79.0 & 83.8 \\
    Transfuser~\cite{transfuser} & C \& L & CM & 97.7 & 92.8 & 92.8 & \textbf{100} & 79.2 & 84.0 \\
    Transfuser{\(^\dagger\)} & C \& L & SNG & 97.8 & 94.5 & 93.5 & \textbf{100} & 80.0 & 85.5 \\
    DRAMA~\cite{yuan2024drama} & C \& L & CM & 98.0 & 93.1 & \textbf{94.8} & \textbf{100} & 80.1 & 85.5 \\
    Hydra-MDP~\cite{hydramdp} & C \& L & CM & 98.3 & 96.0 & 94.6 & \textbf{100} & 78.7 & 86.5 \\
    DiffusionDrive~\cite{diffusiondrive} & C \& L & CM & 98.2 & 96.2 & 94.7 & \textbf{100} & 82.2 & 88.1 \\
    \cmidrule(lr){1-9}
    \textbf{SNG-VLA} & C-single & SNG & \textbf{98.9} & \textbf{96.5} & 92.9 & \textbf{100} & \textbf{83.8} & \textbf{88.24} \\
    SNG-VLA-QA & C-single & SNG & 98.4 & 96.7 & 93.1 & 100 & 83.4 & 88.21 \\
    \bottomrule[1pt]  
    \end{tabular}%
    \egroup
    \caption{\textbf{Comparison on NAVSIM navtest split with closed-loop metrics}. \(\dagger\) represents the results with SNG input. CM represents Driving Command. C-single represents front view image. PDM score (PDMS)\cite{navsim} is weighted aggregation of several sub-scores: no at-fault collisions (NC), drivable area compliance (DAC), time-to-collision (TTC), comfort (Comf.), and ego progress (EP).}
    \vspace{-2mm}  
    \label{tab:navsim_result}
\end{table*}

\begin{table*}[t]
    \centering
    \footnotesize
    \bgroup
    \renewcommand{\arraystretch}{1}
    \resizebox{1\textwidth}{!}{  
    \renewcommand{\arraystretch}{1.1}
    \begin{tabular}{l|c|cccc|c}
    \toprule[1pt] 
    \multirow{2}{*}{\textbf{Method}} & \multicolumn{1}{c|}{\textbf{Open-loop Metric}} & \multicolumn{4}{c}{\textbf{Closed-loop Metric}} & \multirow{2}{*}{\textbf{Latency}} \\
    \cmidrule(r){2-6}
     & Avg. L2 $\downarrow$ & Driving Score $\uparrow$ & Success Rate (\%) $\uparrow$ & Efficiency $\uparrow$ & Comfortness $\uparrow$ \\
    \cmidrule(lr){1-7}
    AD-MLP~\cite{ad-mlp}                & 3.64  & 18.05  &  0.00  &  48.45  & 22.63 & 3ms  \\
    UniAD-Tiny~\cite{uniad}            & 0.80  & 40.73  & 13.18  & 123.92  & \textbf{47.04} & 420.4ms  \\
    UniAD-Base~\cite{uniad}            & 0.73  & 45.81  & 16.36  & 129.21  & 43.58 & 663.4ms \\
    VAD~\cite{vad}                   & 0.91  & 42.35  & 15.00  & 157.94  & 46.01 & 278.3ms \\
    DriveTransformer-Large~\cite{jia2025drivetransformer}                 & \textbf{0.62}  & 63.46  & 35.01  & 100.64  & 20.78 & 211.7ms \\
    \textbf{SNG-VLA}          & 0.82  & \textbf{67.17}  & \textbf{35.90}  & \textbf{158.58}  & 22.30 & 159.6ms \\
    \cmidrule(lr){1-7}
    TCP*~\cite{tcp}                  & 1.70  & 40.70  & 15.00  &  54.26  & 47.80 & 83ms  \\
    TCP-ctrl*~\cite{tcp}             & --    & 30.47  &  7.27  &  55.97  & \textbf{51.51} & 83ms \\
    TCP-traj*~\cite{tcp}             & 1.70  & 59.90  & 30.00  &  76.54  & 18.08 &83ms \\
    TCP-traj w/o distillation~\cite{tcp}  & 1.96  & 49.30  & 20.45  &  \textbf{78.78}  & 22.96 & 83ms \\
    ThinkTwice*~\cite{thinktwice}           & 0.95  & 62.44  & 31.23  &  69.33  & 16.22 & 762ms \\
    DriveAdapter*~\cite{driveadapter}         & 1.01  & \textbf{64.22}  & \textbf{33.08}  &  70.22  & 16.01 &931ms \\
    \bottomrule[1pt] 
    \end{tabular}
    }
    \egroup
    \caption{\textbf{Open-loop and Closed-loop Results in Bench2Drive}. All results are trained on the base training set. Avg. L2 is averaged over the predictions in 2 seconds under 2Hz. * denotes expert feature distillation. Latency is measured on the A6000.}
    \vspace{-2mm}  
    \label{tab:carla_result_1}
\end{table*}

\begin{table}[t!]  
\setlength{\tabcolsep}{5pt}
    \centering
    \small  
    \bgroup
    \renewcommand{\arraystretch}{1.1}  
    \begin{tabular}{l|cccccc}  
        \toprule[1pt] 
        \multirow{2}{*}{\textbf{Method}} & \multicolumn{6}{c}{\textbf{Ability $\uparrow$}} \\  
        \cmidrule(r){2-7}  
        & M & O & EB & GW & TS & Mean \\
        \cmidrule(r){1-7}  
        AD-MLP\cite{ad-mlp}            & 0.0  & 0.0   & 0.0   & 0.0   & 4.4 & 0.9 \\
        UniAD-Tiny~\cite{uniad}         & 8.9  & 9.3   & 20.0  & 20.0  &15.4  & 14.7  \\
        UniAD-Base~\cite{uniad}         & 14.1  & 17.8  & 21.7  & 10.0 & 14.2 & 15.6  \\
        VAD~\cite{vad}                & 8.1  & 24.4  & 18.6  & 20.0 & 19.2 & 18.1  \\
        DriveTransformer~\cite{jia2025drivetransformer}                & 17.6  & \textbf{35.0}  & \textbf{48.4}  & 40.0 & \textbf{52.1} & \textbf{38.6}  \\
        \textbf{SNG-VLA}           & \textbf{33.8} & 11.1 & 46.6 & \textbf{50.0} & 50.0 & 38.1 \\
        \cmidrule(r){1-7}
        TCP*~\cite{tcp}                & 16.12  & 20.0  & 20.0  & 10.0  & 7.0 & 14.6  \\
        TCP-ctrl*~\cite{tcp}              & 10.3  & 4.4  & 10.0   & 10.0  & 6.5 & 8.2 \\
        TCP-traj*~\cite{tcp}              & 8.9  & 24.3  & \textbf{51.7}  & 40.0  & 46.3 & 34.2 \\
        ThinkTwice*~\cite{thinktwice}       & 27.4  & 18.4  & 35.8  & \textbf{50.0}  & 54.2 & 37.2 \\
        DriveAdapter*~\cite{driveadapter}     & \textbf{28.8}  & \textbf{26.4}  & 48.8  & \textbf{50.0}  & \textbf{56.4} & \textbf{42.1} \\
        \bottomrule[1pt] 
    \end{tabular}
    \egroup
    \caption{\textbf{Multi-Ability Results in Bench2Drive}. All results are trained on the base training set. * denotes expert feature distillation.M: Merging, O: Overtaking, EB: Emergency Brake, GW: Give Way, TS: Traffic Sign, M: Mean.}
    \vspace{-2mm}  
    \label{tab:carla_result_2}  
\end{table}

\subsection{SNG-VLA}
\label{sec:architrcture}
We employ LLaVA~\cite{llava} as the backbone, integrating a large language model, Qwen2.5\cite{qwen2}, and a vision encoder, SigLIP~\cite{siglip}. Following the method~\cite{blip2}, we incorporate additional encoders specifically designed for the navigation path and ego state, thereby augmenting the model's capacity to process multimodal inputs. Besides, our model achieves adequate real-time performance.

\subsubsection{Scene representation}
\label{sec:scene_rep}
For a driving scenario, the input is represented as TBT information $I$, navigation path $P=\{(\hat{x}_1, \hat{y}_1), (\hat{x}_2, \hat{y}_2),\dots,(\hat{x}_{N_P}, \hat{y}_{N_p})\}$, front view images $V=(M^1, M^2, ...M^{N_M})$, and the vehicle's ego state $S_t=(v_x^t,v_y^t,a_x^t,a_y^t)$. 
The TBT information $I$ is encoded into a $F_T\in \mathbb{R}^{N_T\times H}$ through LLM tokenizer, where $N_T$ denotes the number of text tokens and $H$ corresponds to the feature dimension of the LLM's transformer backbone. 
Similarly, $P$ is encoded into $F_P\in\mathbb{R}^{N_P\times H}$ through multi-layer perception (MLP) layers. 
We use a pre-trained SigLIP vision encoder~\cite{siglip} to extract features $F_M'$ from multi-view images, which are then projected into $F_M\in \mathbb{R}^{N'\times H}$ via a linear transformation. To mitigate overfitting on ego state inputs\cite{ad-mlp, bevplaner}, we adopt an attention-based state dropout encoder (SDE) inspired by~\cite{plantf}, which applies dropout to each state channel and processes ego state to $F_E\in\mathbb{R}^{4\times H}$.

\subsubsection{Transformer Decoder}
\label{sec:transformer_decoder}
After obtaining the driving scenario representations, all features are concatenated with the waypoint query $Q_W$ into $F$ and fed as input to the transformer backbone. Following the Simlingo~\cite{renz2025simlingo}, the model first auto-regressively generates language predictions that represent the response to the task prompt. Subsequently, in an additional forward pass, it generates waypoint hidden states.

\begin{equation}
F = \text{Concat}(F_T, F_P, F_M, F_E, Q_W)
\label{eq:command_vector}
\end{equation}

The hidden states output by the transformer backbone interact with the navigation query through a cross-attention module, followed by MLP layers to predict the trajectory. The loss function consists of the L1 loss between the predicted and the ground truth trajectory.

\begin{equation}
\mathcal{L} = \| \hat{\tau} - \tau \|
\label{eq:command_vector}
\end{equation}

where the $\hat{\tau}$ denotes the predicted trajectory and $\tau$ denotes the future ground truth trajectory.

\section{Experiments}
\label{experiments}
We evaluate our method in Bench2Drive~\cite{bench2drive}, a closed-loop evaluation benchmark under CARLA Leaderboard 2.0~\cite{carla} for end-to-end autonomous driving. The base set, consisting of 1,000 clips, is used for training, while the model is evaluated on 220 official routes. Additionally, we conduct closed-loop experiments in the NAVSIM benchmark~\cite{navsim} to evaluate its performance in real-world scenarios.

\subsection{Implementation Details}
We employ pre-trained Qwen2.5-0.5B as the transformer backbone and pre-trained SigLIP-So400M as the vision encoder, with a patch size of 14 and an image size of 384. In the state dropout encoder (SDE), we apply a dropout rate of 0.5 to the four ego state channels. The visual inputs consist of front and rear camera images, which undergo additional downsampling after passing through the vision encoder. The SNG-QA task is employed exclusively in the NAVSIM experiment. Due to real-time constraints in CARLA and the availability of detailed scenario labels, we generate SNGs through rule-based methods in our Bench2Drive experiments without employing SNG-QA. We use a learning rate of 1e-6 with a cosine annealing schedule and a warmup ratio of 0.03. The model is trained on 8 $\times$ NVIDIA A100 GPU 80G with a per-GPU batch size of 8 for 10 epochs.

\subsection{Main results}

We compare our method with other E2E-AD methods in both Bench2Drive and NAVSIM. Table~\ref{tab:navsim_result} presents the results on NAVSIM~\cite{navsim}. 
After adding SNG, Transfuser's performance has been significantly enhanced, with improvements observed in both DAC and TTC. The performance of Hydra-MDP~\cite{hydramdp} is enhanced through additional training to optimize for the EP evaluation metric, utilizing supplementary supervision and weighted confidence post-processing. Despite this, our model achieves SOTA performance without relying on any supervision from perception tasks. Our model exhibits improvement on the DAC (drivable area compliance) metric, which underscores the enhanced efficacy of the proposed SNG. The SNG-VLA-QA model can perform reasoning tasks while maintaining planning performance.

\begin{figure*}[t!]
    \centering
    \includegraphics[width=\textwidth]{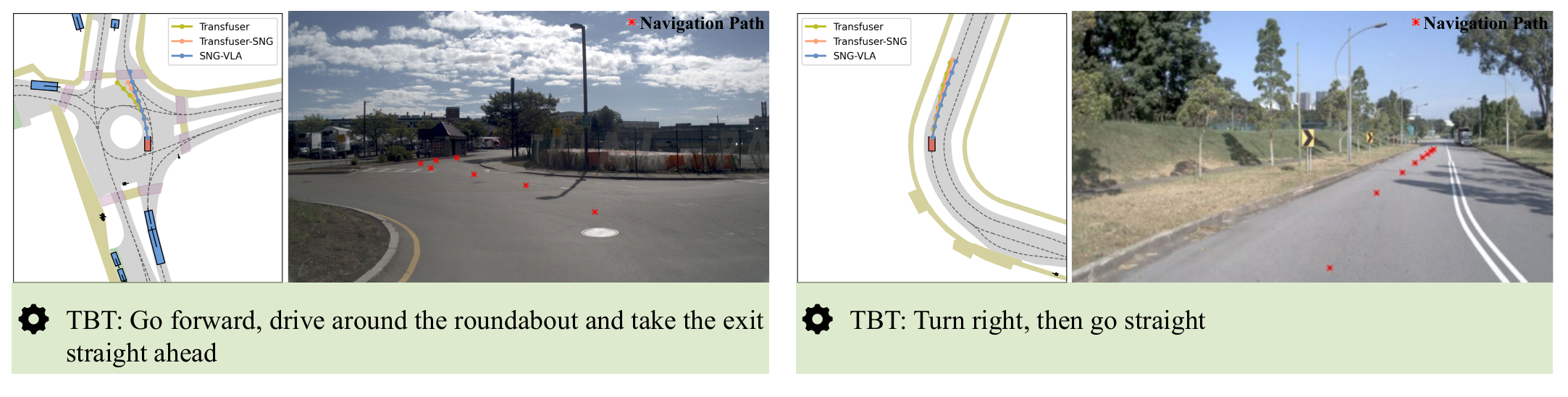}
    \caption{
    Qualitative analysis of real-world scenarios. Navigation paths are augmented with substantial noise before being fed into the model to mitigate the influence of privileged information.
    }
    \vspace{-2mm}  
    \label{fig:qualitative}
\end{figure*}

\begin{figure}[t!]
    \centering
    \resizebox{0.5\textwidth}{!}{  
    \includegraphics[width=0.5\textwidth]{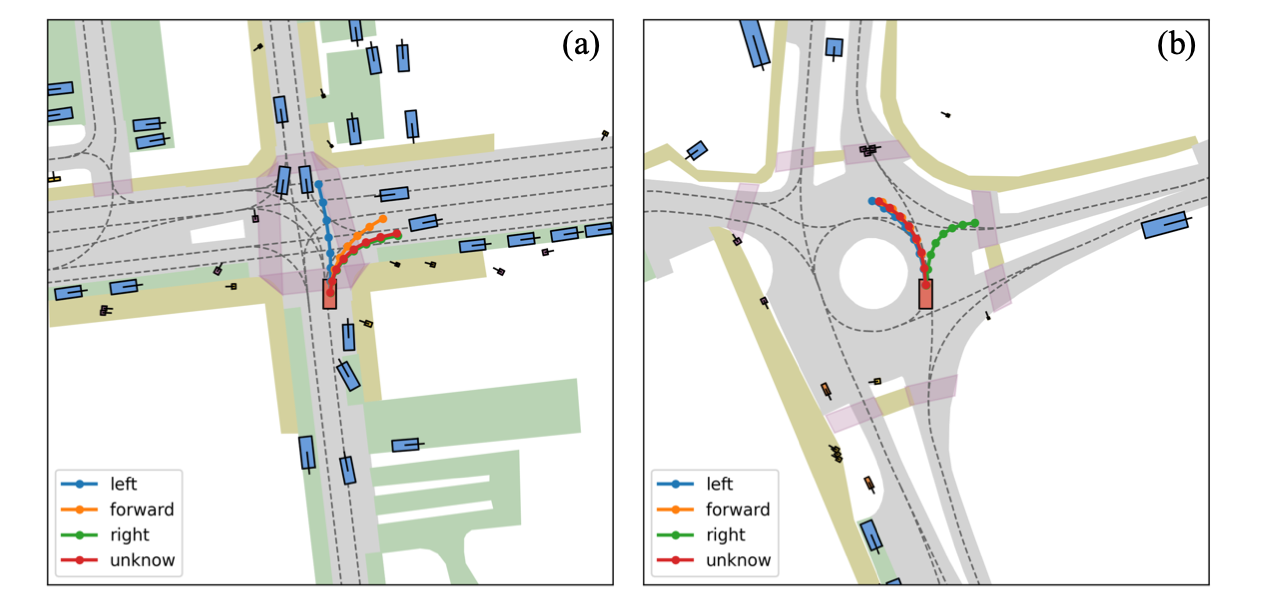}
    }
    \caption{
    We demonstrate the impact of introducing noise to the driving command on the predicted trajectory during the inference process of the Transfuser~\cite{transfuser}. The experimental results are based on the official checkpoint~\cite{navsim}.
    }
    \vspace{-2mm}  
    \label{fig:cmd_ablation}
\end{figure}

Table~\ref{tab:carla_result_1} and Table~\ref{tab:carla_result_2} present the results in Bench2Drive. Models based on driving commands exhibit insufficient modeling of navigation information, leading to target loss during trajectory planning. Consequently, these models generate trajectories that deviate randomly from the intended path. Furthermore, cumulative errors in closed-loop experiments further degrade the performance of driving command-based methods, leading to poor performance in key metrics such as task completion rate and driving score. In contrast, our SNG-based model has significantly outperformed existing methods.
Compared to UniAD-Base~\cite{uniad}, our method surpasses it by 46.6\% and 119.4\% in terms of Driving Score and Success Rate, respectively. In the mean Multi-Ability score, our method outperforms VAD~\cite{vad} by 110.7\%. We employed a smaller backbone with a single vision encoder that processes only the front view image, achieving favorable latency performance without text generation tasks.

\subsection{Qualitative Results}
Fig.~\ref{fig:qualitative} presents the qualitative experimental results of our model. In both scenarios, transfuser-sng demonstrates superior performance compared to transfuser. The incorporation of SNG enables the model to more effectively comprehend global navigation information and subsequently predict the planning trajectory in accordance with the navigation guidance. Similarly, SNG-VLA exhibits remarkable capabilities in complex scene understanding and navigation following.

\subsection{Ablation study}

\subsubsection{Driving command fails to model navigation information}

Undoubtedly, clear navigation information plays an important role in autonomous driving systems, providing essential guidance for determining the vehicle's direction of driving. However, as shown in Table~\ref{tab:cmd_ablation}, after introducing varying levels of noise into the driving command, the model can yield results that are comparable to, or surpass the official results on metrics on PDM score and others. The use of random or fixed driving commands has minimal impact on model performance.
We present qualitative results in Fig.~\ref{fig:cmd_ablation}. 
In an open intersection scenario where the expert trajectory demonstrates a right turn, the model fails to execute ``Go Forward" and ``Turn Left" commands. In the roundabout scenario where the expert trajectory performs a left turn, when given a ``Turn Right" command, the model generates a counter-flow right turn trajectory that could cause severe accidents. Notably, in both scenarios, when the driving command is set to ``Unknown" the model's output performance is comparable to that achieved with the correct driving commands.

\subsubsection{Driving command vs. Sequential navigation guidance}
As illustrated in Table~\ref{tab:cmd_vs_SNG} (ID 0-4), the results obtained without any navigation information (ID 0) and with driving commands (ID 1) exhibit no significant difference. The performance achieved by solely utilizing TBT information (ID 2) surpasses that of both (ID 0) and (ID 1). Notably, when employing only two points spaced 20 meters apart as the navigation path (ID 3), the model's performance is comparable to that of (ID 2), indicating that the navigation path can provide the model with effective spatial reference under sparse sampling. The model achieves its optimal performance when both the navigation path and TBT information are used as sequential navigation guidance (ID 4). These results demonstrate that sequential navigation guidance models navigation information more effectively than driving commands alone. Specifically, the navigation path provides long-term trajectory constraints, while TBT information offers real-time decision-making logic, such as road traffic conditions. The synergy between these two elements mitigates the limitations associated with relying on a single modality.

\begin{table}[t!]
    \centering
    \footnotesize
    \bgroup
    \renewcommand{\arraystretch}{1.1}
    \begin{tabular}{l|cc|ccc|c}
    \toprule
    Command & NC$\uparrow$ & DAC$\uparrow$ & TTC$\uparrow$ & Comf.$\uparrow$ & EP$\uparrow$ & PDMS$\uparrow$ \\
    \cmidrule(lr){1-7}
    Original   & 97.7 & 92.8 & 92.8 & 100 & 79.2 & 84.0 \\
    None    & 97.7 & 93.5 & 92.9 & 100 & 78.7 & 84.4 \\
    Random  & 97.8 & 93.4 & 93.3 & 100 & 79.2 & 84.7 \\
    Left    & 97.4 & 93.4 & 92.5 & 100 & 79.2 & 84.3 \\
    Right   & 97.7 & 93.5 & 93.2 & 100 & 78.9 & 84.4 \\
    Forward & 97.7 & 93.2 & 93.1 & 100 & 78.9 & 84.2 \\
    \bottomrule
    \end{tabular}%
    \egroup
    \caption{\textbf{Command ablation results in NAVSIM}. We conducted experiments on NAVSIM~\cite{navsim} using Transfuser~\cite{transfuser} by modifying the input driving commands. Original: ground truth driving command; None: no driving command added; Random: random driving command; Left, Right, Forward: fixed driving commands. All results are obtained under identical training parameters. The introduction of noise into the driving command has minimal impact on the final planning performance.
    }
    \vspace{-2mm}  
    \label{tab:cmd_ablation}
\end{table}

\begin{table}[t!]
\setlength{\tabcolsep}{5pt}
    \centering
    \footnotesize
    \bgroup
    {  
    \renewcommand{\arraystretch}{1.1}
    \begin{tabular}{c|c|c|c|cc|c}
    \toprule
    ID & \makecell{Navigation \\ Path} & \makecell{TBT \\ Information} & \makecell{Driving \\ Command} & NC$\uparrow$ & DAC$\uparrow$ & PDMS$\uparrow$ \\
    \cmidrule(lr){1-7}
    0 & \textminus & \textminus & \textminus & 97.2 & 95.1 & 85.9 \\
    1 & \textminus & \textminus & \checkmark & 97.5 & 95.3 & 86.1 \\
    2 & \textminus & \checkmark & \textminus & 97.6 & 95.2 & 86.4 \\
    3 & $2\times20$ & \textminus & \textminus & 97.8 & 95.1 & 86.4 \\
    4 & $2\times20$ & \checkmark & \textminus & 97.5 & 96.1 & 87.6 \\
    5 & $4\times10$ & \textminus & \textminus & 97.5 & \textbf{96.6} & 87.7 \\
    6 & $4\times10$ & \checkmark & \textminus & \textbf{98.9} & 96.5 & \textbf{88.2} \\
    7 & $8\times5$ & \textminus & \textminus & 97.5 & 96.2 & 87.2 \\
    8 & $8\times5$ & \checkmark & \textminus & 97.5 & 96.6 & 87.6 \\
    \bottomrule
    \end{tabular}%
    }
    \egroup
    \caption{\textbf{Ablation of navigation information representation.} We conduct ablation studies on the sampling interval of the navigation path, TBT information and driving commands. }
    \vspace{-2mm}  
    \label{tab:cmd_vs_SNG}
\end{table}


\subsubsection{Optimal combination of navigation path \& TBT information}
We further investigated the optimal combination of navigation paths and TBT information, as practical operations often face challenges in consistently obtaining sufficiently dense navigation paths or accurate TBT information. As shown in Table~\ref{tab:cmd_vs_SNG} (ID 3-8), we systematically evaluated the impact of varying densities of navigation path points and the inclusion of TBT information on the results. Comparing (ID 3, 5, 7), it's shown that 4 navigation points spaced 10 meters apart yielded the best performance. Both excessively sparse and overly dense configurations led to diminished performance. Sparse navigation points fail to accurately model the reference path, while overly dense points impose excessive constraints on the model, resulting in poor performance in scenarios such as obstacle avoidance. Across (ID 3-8), the inclusion of TBT information consistently improved performance under varying navigation point densities. In conclusion, a moderate-density navigation path combined with TBT information serves as an effective SNG setting, optimally modeling navigation information and maximizing the model's planning performance.

\begin{figure}[t!]
    \centering
    \includegraphics[width=0.5\textwidth]{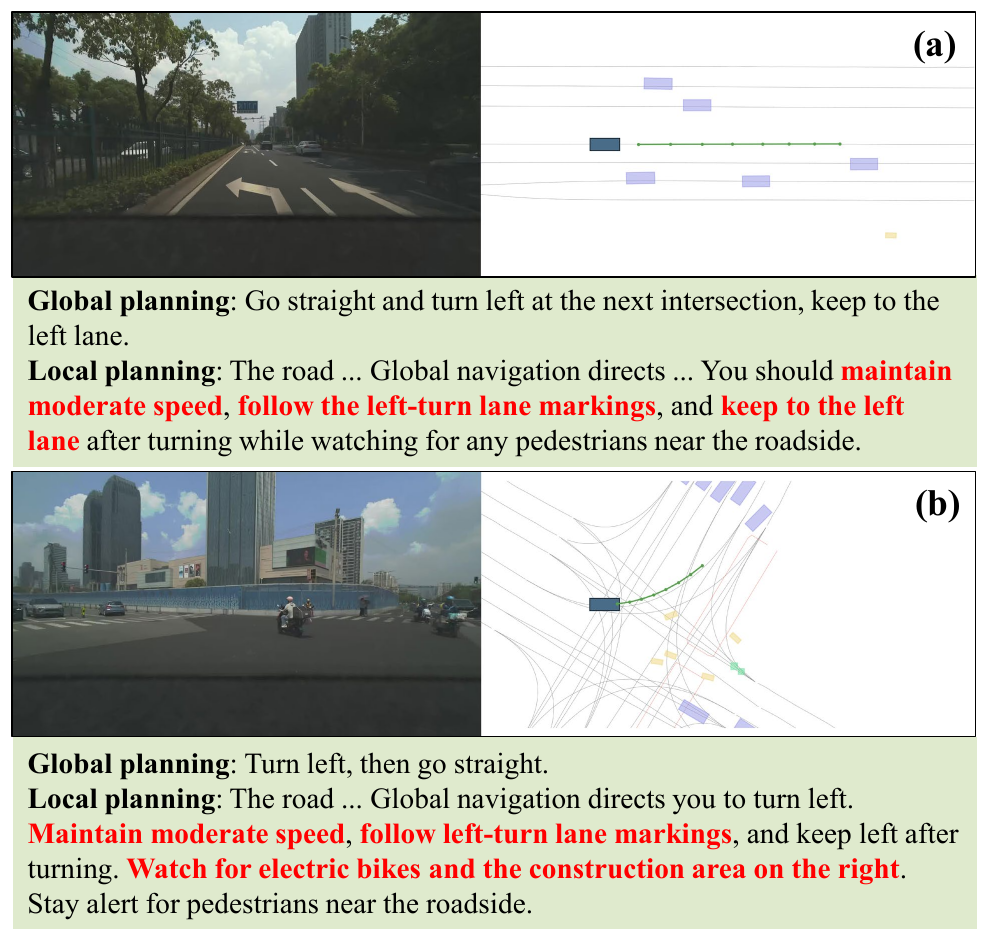}
    \caption{
    Qualitative analysis of real-world scenarios. 
    }
    \vspace{-2mm}  
    \label{fig:real_world}
\end{figure}

\subsection{Real world experiments}



To further evaluate our proposed SNG and SNG-VLA in real-world scenarios, we established a validation platform. The platform is equipped with a LiDAR, the Innovusion Falcon 300, and a surround-view camera system comprising five AR0820 cameras with a 120-degree horizontal field of view (HFOV) and two AR0820 cameras with a 70-degree HFOV. The onboard computing unit consists of dual Orin modules. We collect an in-house dataset and conduct experiments. We present the model's reasoning process and planned trajectories in Fig.~\ref{fig:real_world}. Our method performs excellently in selecting turning lanes and providing warnings for special scenarios, pedestrians, and electric vehicles.

\section{Conclusions}
\label{conclusions}
In this study, we investigate the limitations of end-to-end autonomous driving systems in utilizing navigation information and introduce a novel representation, termed Sequential Navigation Guidance, which integrates long-term trajectory constraints and real-time decision logic. Our model SNG-VLA, achieves superior performance in closed-loop evaluations without the supervision from perception tasks. Experiments conducted in real-world scenarios further confirm the robustness and practical applicability of our approach.




\bibliographystyle{IEEEtran}
\bibliography{references} 

\end{document}